\documentclass[10pt,a4paper]{article}
\usepackage[nonatbib, preprint]{neurips_2019}
\usepackage{lipsum}
\usepackage[export]{adjustbox}
\usepackage{cuted}
\usepackage{enumitem,amsmath,amssymb,graphicx,float}
\usepackage[sort,square]{natbib}
\usepackage[utf8]{inputenc} 
\usepackage[T1]{fontenc}    
\usepackage{hyperref}       
\usepackage{url}            
\usepackage{booktabs}       
\usepackage{amsfonts}       
\usepackage{nicefrac}       
\usepackage{microtype} 
\usepackage{subfigure}
\expandafter\def\csname ver@subfig.sty\endcsname{}
\usepackage{subfig}

\title{Sepsis World Model: A MIMIC-based OpenAI Gym "World Model" Simulator for Sepsis Treatment}

\author{%
  Amirhossein Kiani\\
  Dept. of Computer Science\\
  Stanford University\\
  \texttt{kiani@cs.stanford.edu} \\
  \And
  Chris Wang\\
  Dept. of Computer Science\\
  Stanford University\\
  \texttt{chrwang@stanford.edu} \\
  \And
  Angela Xu\\
  Dept. of Computer Science\\
  Stanford University\\
  \texttt{aaxu@stanford.edu} 
}

\begin{document}

\maketitle

\vspace{-40pt}
\vspace{5 pt}

\section {Introduction and Motivation}
\vspace{-5 pt}
Sepsis is a life-threatening condition caused by the body’s response to an infection. In order to treat patients with sepsis, physicians must control varying dosages of various antibiotics, fluids, and vasopressors based on a large number of variables in an emergency setting. With the onset of large digital health record datasets such as the MIMIC dataset \citep{mimic-dataset}, machine learning is an increasingly popular approach used for sepsis outcome and treatment prediction. MIMIC is a large, single-center database consisting of the information relating to patients admitted to critical care units at a large tertiary care hospital and includes a rich collection of information such as patient demographics, vital signs, labs, medical procedures and survival data. Using such datasets, deep reinforcement learning has been applied to the task of learning optimal policies for sepsis treatment in works such as \cite{SepsisRL}. However, one major challenge with applying deep reinforcement learning to learning from EHR datasets is that our known states only consist of a sample of the entire state space, a sample which also contains noise. Existing work has used solutions such as off-policy evaluation with importance sampling, or tried training stochastic policies and other evaluation techniques \citep{representation} to overcome this.

In this project we employ a "world model" \citep{Ha2018} methodology to create a simulator that aims to predict the next state of a patient given a current state and treatment action. In doing so, we hope our simulator learns from a latent and less noisy representation of the EHR data. Using historical sepsis patient records from the MIMIC dataset, our method creates an OpenAI Gym simulator that leverages a Variational Auto-Encoder and a Mixture Density Network combined with a RNN (MDN-RNN) \citep{Ha2018} to model the trajectory of any sepsis patient in the hospital. To reduce the effects of noise, we will sample from a generated distribution of next steps during simulation and have the option of introducing uncertainty into our simulator by controlling the "temperature" variable similar to \citet{Ha2018}. It is worth noting that we do not have access to the ground truth for the best policy because we can only evaluate learned policies by real-world experimentation or expert feedback. Instead, we aim to study our simulator model's performance by evaluating the similarity between our environment's rollouts with the real EHR data and assessing its viability for learning a realistic policy for sepsis treatment using Deep Q-Learning.

\vspace{-5 pt}
\section{Approach}
\vspace{-5 pt}
\subsection{Dataset Overview and Preprocessing} \label{sec_dataset}
\vspace{-5 pt}

 To construct the necessary models, we use the MIMIC dataset. This large and comprehensive dataset consists of the health trajectories of 40,000 critical care patients during their hospital stays. Among the sepsis patients, our preprocessed dataset includes individual patient datapoints over time such as patient demographics, vital signs, laboratory tests, medications, medical interventions, and outcome. A concrete sample of our already processed data is located in the Appendix section.

Each state consists of 46 normalized features from the dataset.  Possible actions are discrete numbers between 0 and 24 indicating the space of possible vasopressor and IV fluid interventions across 5 dosage quantiles. The end goal is to leverage this dataset to be able to suggest a treatment action for each time step based on the information known about any patient at a particular time step, with the objective of ensuring that the patient survives. In order to achieve this we build a "State Model" to predict the next state of a patient given its current state and performed action, which is described in the next section. 

\subsection{Simulator Models} 
\vspace{-5 pt}
Our baseline consists of three standard RNN models that simulate next state (state model), end of stay (termination model) and outcome prediction (outcome model). These models were built through a course project done by one of the project members\footnote{\url{https://github.com/akiani/rlsepsis234}}. The baseline does not model the uncertainty of the states using a MDN-RNN and does not leverage VAEs, which results in a simulator that overfits to noisy datapoints. 

\begin{figure}[h]
\centering 
\begin{minipage}{.4\textwidth}
\includegraphics[width=\linewidth]{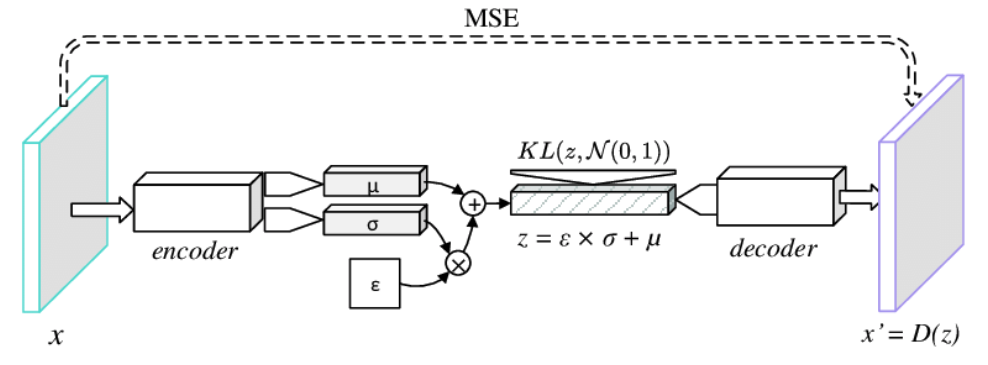}
	\caption{A Standard Variational Auto-encoder by \citeauthor{Pinho}}
	\label{vae_fig}
\end{minipage}	
\quad \quad
\begin{minipage}{.4\textwidth}
	\includegraphics[width=\linewidth]{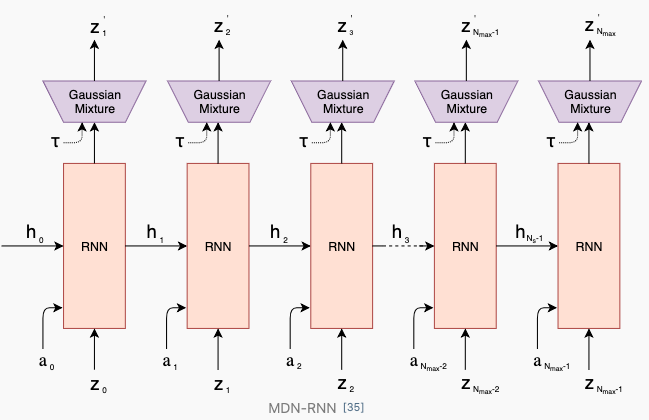}
	\caption{RNN with a Mixture Density Network output layer. The MDN outputs the parameters of a mixture of Gaussian distribution used to sample a prediction of the next latent vector $z$.}
	\label{MDN}
\end{minipage}
\end{figure}

The new simulator model we build for this project consists of two components: a Variational Auto-encoder (VAE) and MDN-RNN. As shown in Figure \ref{vae_fig}, the VAE takes in the noisy patient states each of 46 features and encodes them into a smaller, more compact latent state representation $z$ of 30 features by sampling from the learned probability distribution parameters ($\mu$ and $\sigma$).  We implemented the VAE using Tensorflow, with three dense downsampling layers in the encoder, sampling via reparameterization, and three upsampling layers in the decoder. The latent dimension of 30 was determined upon experimentation. We minimized the mean-squared error (MSE) between the the input and reconstructed output created by the decoder. 

 At each timestep $t$, we use the trained VAE model to pre-process and denoise the state data point to produce the encoding $z_t$  which is fed as input to the MDN-RNN model.  The stochastic MDN-RNN model, as shown in Figure~\ref{MDN}, predicts the probability distribution of the next state $z_{t+1}$ in the next time step as a Mixture of Gaussian distributions conditioned on $a_t$ and $h_t$, the current action and hidden state at timestep $t$, respectively.  At every timestep in the simulation, we sample from this distribution of possible next state features, that is, $p(z_{t+1} | a_t, z_t, h_t)$.  The motivation behind building a stochastic model for the patient state space is to account for uncertainties in the state feature space.
 
 This simulator which combines both the VAE and MDN-RNN is inspired from the "World Models" paper by \citeauthor{world-model} and is shown in Figure \ref{world_sim_diagram}. We perform two kinds of analyses on the simulation environment which is shown in the center of the diagram.  On the left side, we train a DQN provided by Open AI Baseline to learn the optimal policy.  On the right side, we simulate the physician's policy's rollout.  Both of these methods are described further in the Simulator and Evaluation sections.

\begin{figure}[h]
\centering 
\includegraphics[width=0.5\textwidth]{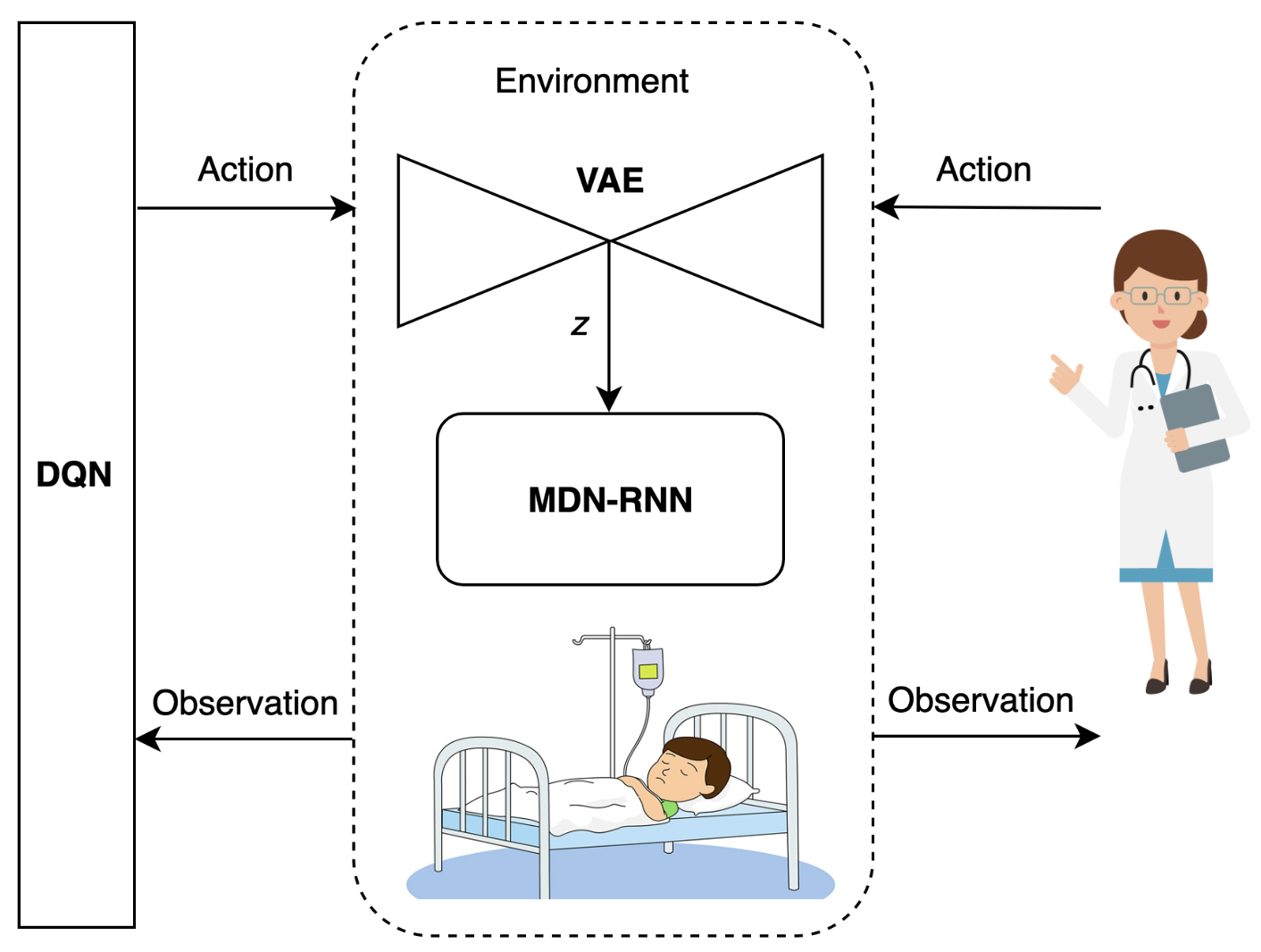}
	\caption{Our Patient "World Model"}
	\label{world_sim_diagram}
\end{figure}

After the VAE produces the latent states, we feed these encodings into the aforementioned RNN models (state model, termination model, and outcome model) described in greater detail below.  We experimented with combinations of VAE + RNN, MDN + RNN, and VAE + MDN + RNN to see if we could improve the RNN baseline, and to analyze which improvements were helpful to the baseline. 

\vspace{-5pt}
\subsection{State Model}
\begin{figure}[h]
\centering 
\includegraphics[width=0.32\linewidth]{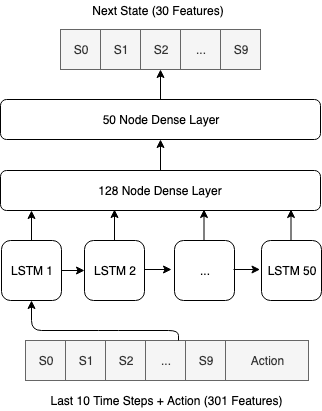} \ \ 
\centering
\includegraphics[width=0.32\linewidth]{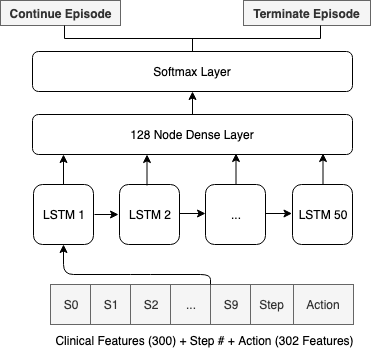} \ \ 
\centering
\includegraphics[width=0.32\linewidth]{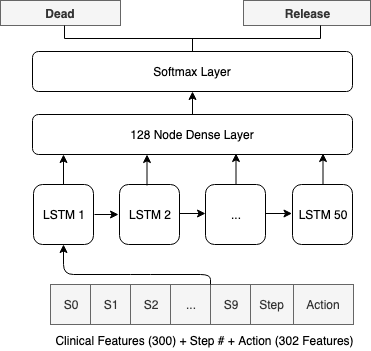}
\centering
\caption{Simulator Model Architectures}
\label{simulator_models}
\end{figure}

The state model is an RNN trained using pre-processed MIMIC features from our training data. The architecture of this model is described in Figure \ref{simulator_models}.  The input to this model consists of the encoded states produced by the VAE encoder in the past and an action value (0-24) for the current time step (zero padded). The output of the model is the features representing the next state.

\vspace{-5pt}
\subsection{Episode Termination Model}
A separate model was developed to detect episode transitions. The transitions were defined as two mutually exclusive cases of (1) terminating the episode (2) continuing the
episode. We accounted for the length of the episodes by
adding an step number feature to the space and action
features for this model. The intuition for this feature was to
model the distribution of episode lengths seen in our training
data. The detailed architecture of the episode termination model is demonstrated in
Figure 2.
\vspace{-5pt}
\subsection{Episode Outcome Model}
A third model with the same features and architecture as the
Episode Termination Model was developed to predict the
two mutually exclusive outcomes of death or release from
hospital. This model was used in the environment to decide
the reward values at the end of each episode.
\vspace{-5pt}
\subsection{DQN Agents Model}
To complete the evaluation of our simulator, we leveraged
OpenAI Baselines off the shelf algorithms to train three
agents on top of our OpenAI gym environments, which  are based  on our different architecture choices (baseline, VAE, MDN, VAE + MDN). We use an OpenAI Gym wrapper with each of our simulators so that given a state, taking a specific action step will yield a new generated state and reward. Additionally, the simulator will use our termination and  outcome models to determine whether the new state is a terminal state and if so, what the reward should be.

We then use the off the shelf OpenAI Baselines \footnote{\url{https://github.com/openai/baselines}} framework, which takes in  a Neural Network Q Function approximator (in  our  case two hidden layers with size 128 and a $tanh$ activation function), as well as an environment, and attempts to learn the optimal policy in the simulation environment. 

Of course, one key variable in this setup is the formulation of the reward function, which significantly impacts the selection of the optimal action. We decide to test out three reward functions for the environment. We further compare the results of our learned policies with the physician's policy across following configurations:
\begin{enumerate}
    \item Only the end-of-episode outcome is taken into consideration, and we assign a reward of $\pm 15$ based on the outcome. 

    \item In  addition to the end-of-episode rewards, we penalize the model for  taking extreme actions; i.e.  we assign a reward of $\pm 1000$ based  on the outcome and then penalize the model  at each step in an amount equal to the negative of the action's intensity.

    \item We assign a reward of $\pm 15$ based  on the outcome and then  use intermediate lactate and SOFA levels to calculate  additional reward at each  step, as follows:
    \begin{figure}[h]
    \centering 
    \includegraphics[width=1\textwidth]{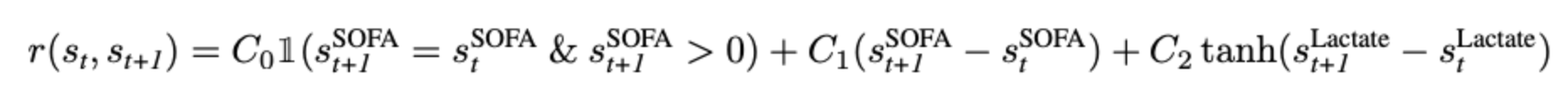}
    	\caption{Reward function  on intermediate steps}
    	\label{world_sim_diagram}
    \end{figure}
    We borrowed this formulation from \citeauthor{SepsisRL}. The motivation behind this function is to simultaneously ensure that the intermediate  rewards do not overshadow the final reward and provide some guiding feedback to move the policy in the right direction at each individual timestep.
\end{enumerate}

\section{Results and Analysis}
\subsection{Autoencoder  and VAE}
We trained several different configurations for our VAE, in which we varied the hyperparameter controlling the proportion of the KL-loss we include (relative to the reconstruction loss) and the size of the hidden layers. Eventually, we decided on an encoding size of 30 dimensions without incorporating KL loss, because we found the KL loss would often diverge inconsistently. We tried two different configurations of layer patterns, one with hidden layers of 45 and 40  and one with hidden layers of 40 and 35. Eventually, we decided to use the latter formulation, as it yielded more realistic curves, which we hypothesize is due to the more balanced reduction of the number of hidden states at each layer. For comparison purposes, we also trained a traditional Auto Encoder (AE) without a Variational component.  

\begin{figure}[h]
\centering 
\includegraphics[width=0.7\textwidth]{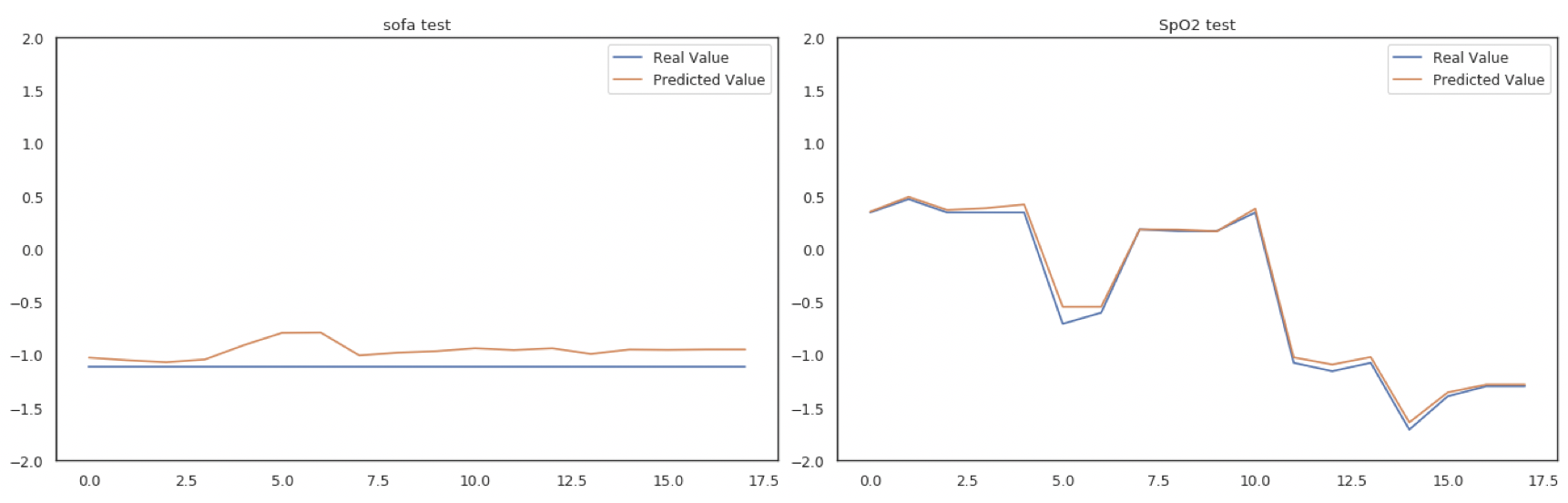}
\centering
\caption{Comparison between VAE decoded states and real states.}
\label{vae_test_plots}
\end{figure}

After training the VAE for 20 epochs, we report a final reconstruction loss (or distance between the reconstructed state and original state) of $0.0791$ for the VAE, just slightly worse than our AE. We visualized the reconstructed states compared with the real trajectories for several of the 46 clinical features and confirmed that our VAE was generating reasonable predictions.  Figure \ref{vae_test_plots} shows comparison plots between the predicted states by the VAE and the real states for clinical features.

Although an analysis of reconstruction loss and similarity plots is important for debugging the VAE, it is also important to be aware that the purpose of using the VAE in this application is to reduce noise and point out salient features in the original data, not necessarily to match all state features.

\subsection{Simulator: State, Termination, and Outcome}
\begin{figure}[h]
\centering 
\includegraphics[width=1\textwidth]{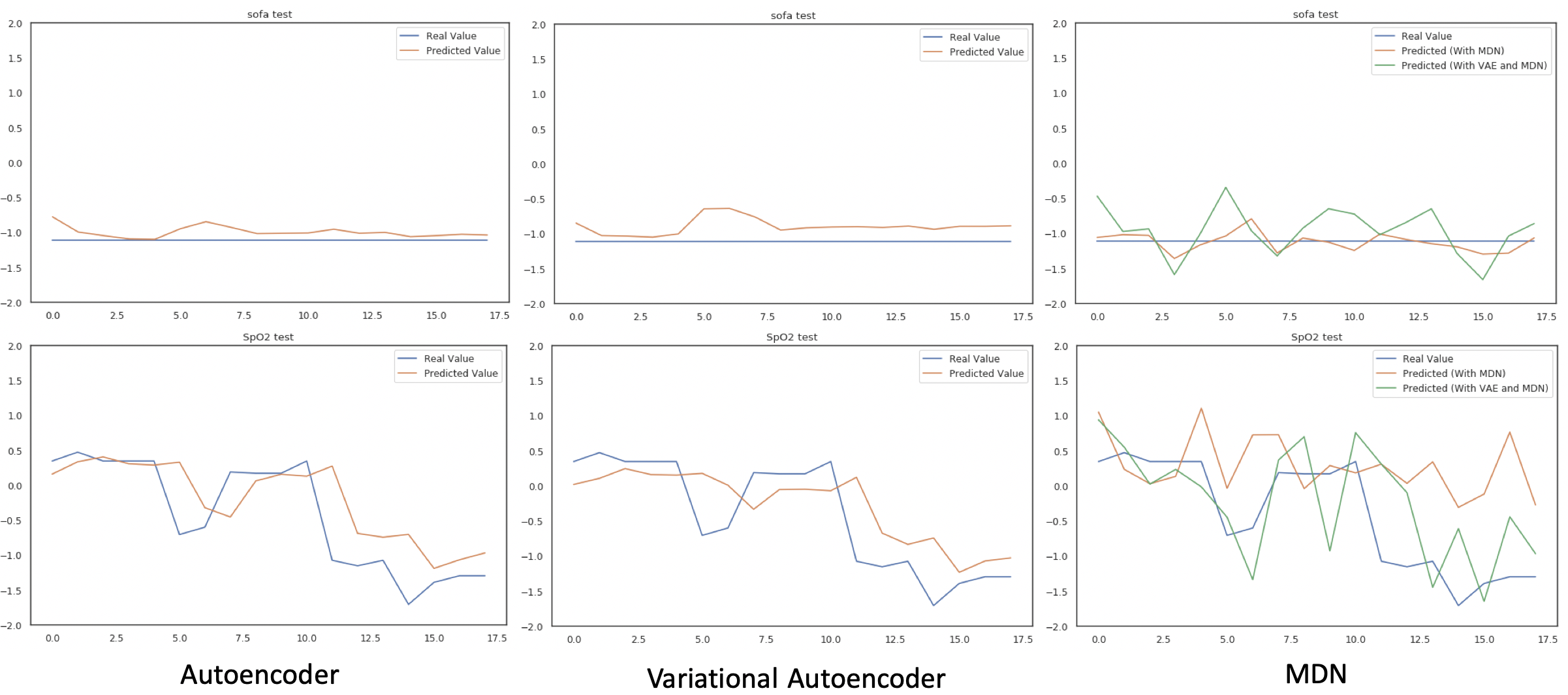}
\caption{Predictive Power of State Simulator}
\label{state_model_plots}
\end{figure}

 Table \ref{simulator_tab} describes the accuracy and loss values for the best chosen models for the state, transition and outcome models. We ran all models using Tensorflow, and used early stopping with a patience value of 3, meaning that if the models validation metric did not improve for 3 epochs, we used the last best parameters based on this metric and automatically terminated the training. All models reached this state within the first 10 epochs. It seems as if the state, episode termination, and outcome prediction models are able to achieve a similar level of accuracy with the VAE  as without, which indicates that while the VAE may be losing state based information, it is still capturing the information necessary to perform downstream tasks, which is a good sign. Note that MDN network is only used for the state model and not the termination and outcome models. Conceptually, since the termination and outcome are boolean values, predicting a mixture of distributions for their values may not be as effective. Despite seemingly good results, these high level loss metrics do not provide us with a clear picture of our state predictions.
\begin{table}[h]
\centering
\includegraphics[width=1\textwidth]{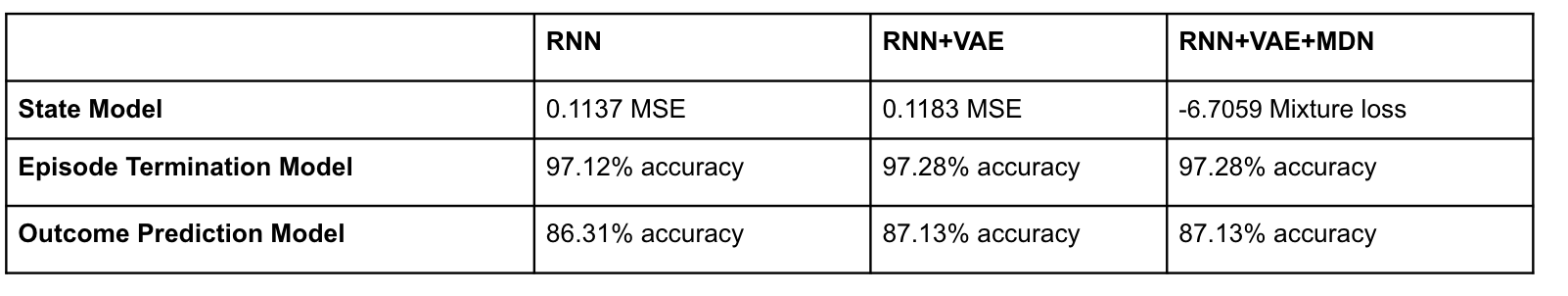}
\caption{Accuracy and loss values for the best chosen models for the state.}
\label{simulator_tab}
\end{table}
Thus, we evaluated the predictive power of our state simulator by feeding the simulator with states from the previous 10 time steps and action values from the real data trajectories and predicted the next states. Figure \ref{state_model_plots} plots the simulated projections against the real values for the SOFA and SpO2 state features across the length of an episode for the different models we experimented with, including the regular Autoencoder, the VAE, the VAE + MDN, and the MDN by itself. Keep in mind that these plots are not a completely realistic simulation (which we will explore in the next section). In these plots even if the model incorrectly predicts a state, it will receive the correct version of  the  state as input for predicting the next step. The results show that the MDN imparts much more variance into the predicted state, as expected.  It seems to be learning more than the AE and VAE models; instead of simply keeping the predictions constant until an old state is added back in as a input (you can sort of see this trend with  the yellow movements happening after the blue on the two left graphs), decisions  seem to be made by the model.  The MDN+VAE has even more variance than the MDN itself, as expected, and seems to detect a general trend  in the SpO2, which looks quite promising. These observations motivate a fully fledged rollout and comparison with the simple RNN, which we do in the next section, in order to verify whether the MDN-RNN models are truly learning something meaningful.


\subsection{Analysis of Rollout on Physician's Policy}
\begin{figure}[h]
\centering 
\includegraphics[width=1\linewidth]{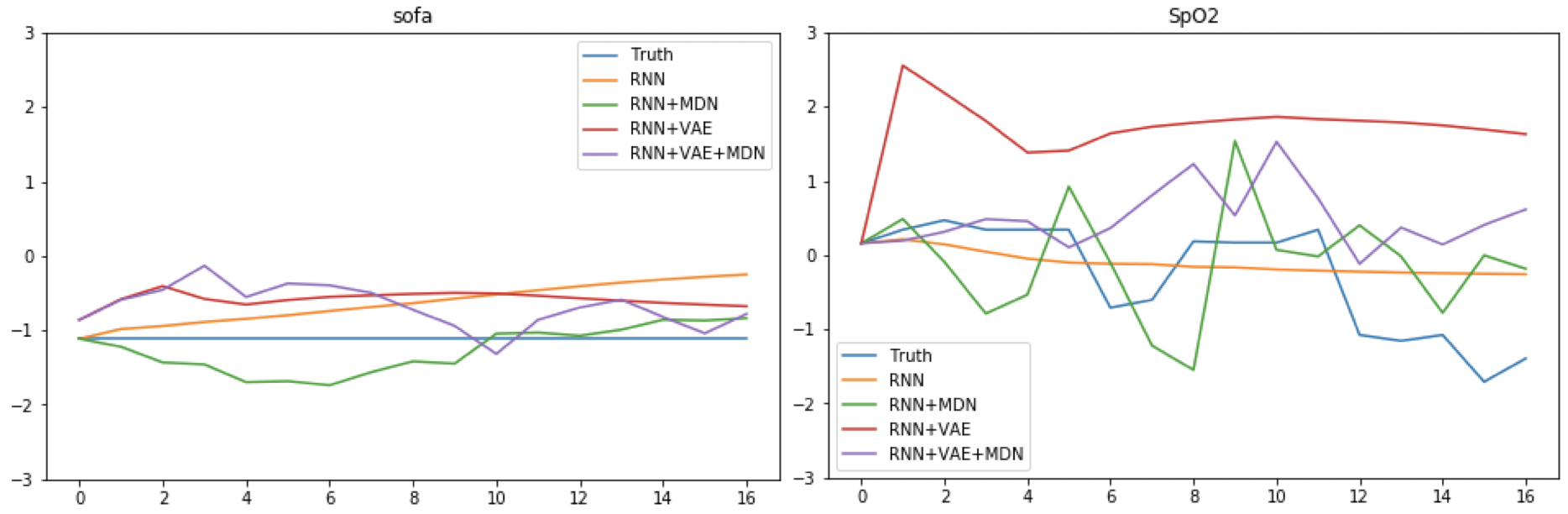}
\centering
\vspace*{-4mm}
\caption{Rollout on Physician's Policy. Comparison of different networks' effects on rollout stability.}
\label{rollout_plots}
\end{figure}

We attempted to visually inspect our simulator on the physician's policy. Specifically, we initialize the model with the starting state of a patient. We then performed the
actual series of actions that the physician performed on each
patient and visualized the state features across the length of
the episodes. Note that compared to the previous section, here we only have access to the states that the model generated itself as the history; this is representative of what we may have access to when trying to train an agent to learn a policy through exploration, as we wont necessarily have access to the infinitely sized and continues state-action space in our dataset.
The results are demonstrated in Figure \ref{rollout_plots}.  

The rollout results indicate that the RNN, by itself, produces smooth curves, as opposed to a constantly varying trend in MDN-based models. We believe this may be due to the fact that the RNN itself was unable to fully capture the dynamic variance of the outputs, and thus converges to finding a "mean"  of the potential  next states. This is the reason we introduced an MDN-RNN in the first place, so that the model could predict a group of distributions that the next  state is from and capture the idea that the state must come from solely one  of them. (As an example, if there was a 50 percent chance a patient's SpO2 shot up to 10 and a 50 percent chance it went to -10, a traditional RNN  would likely predict  it to be 0 to mimimize  MSE. An MDN-RNN would be able to simulate 10 half the time and -10 half the  time).  Indeed, the RNN+MDN seems to better capture the variance across the episode  and follow the general trend--although with  this comes the risk that sometimes the wrong distribution might be chosen, and this will make the predicted state veer off  the real state even further than the  traditional RNN. We notice  that with the MDN, while individual steps may have large variance from the  previous step, the distribution usually corrects itself in the next prediction back to a more stable value (it can be  debated whether this is a good thing). 

In this particular rollout, we notice that the RNN+VAE rollout  has a lot of trouble on the SpO2 prediction, even without a MDN. Perhaps the RNN had trouble with these specific encoded states but keep in  mind the purpose of the VAE is not to perfectly reproduce the original state features. Overall, however, it seems like that the  MDN and VAE are successful in modeling the variance and distributions that the next state can be  drawn from.



\vspace{-5pt}

\subsection{Normalized Trajectory Means}
\begin{figure}[H]
\centering 
\includegraphics[width=1\textwidth]{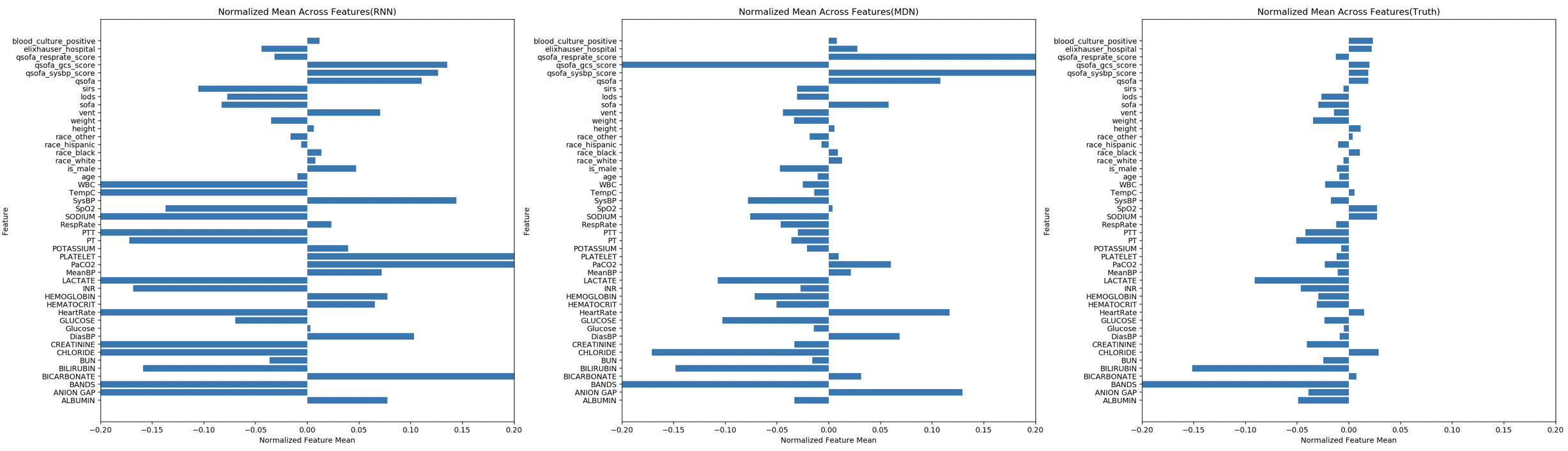}
\centering
\vspace*{-4mm}
\caption{Comparison of Trajectory Normal Means for RNN, MDN, and Truth.}
\label{trajectory_plots}
\end{figure}

We would like to have a quantitative metric in order to measure the compounding error for the results of the simulator as described in the last section. We propose the Normalized Trajectory Mean metric which computes, for each feature, the mean of that feature across all rollouts based on the particular state model. We measured this value for different
features as indicated in \ref{trajectory_plots}. When episodes ended earlier in the simulator compared to the real world or vice versa, we imputed a value of zero for the missing values. We also normalized each features' mean by the sum of squares of the feature values in the real dataset to generate
a comparable weight for each feature. While this metric does not encapsulate  variance as much (which we can analyze by looking at the graphs, as in the above section), it provides an overview of how well calibrated the model is for each metric. It
can act as a sanity check for our model's performance and
provide a direction for prioritizing future improvements.  
\vspace{-5pt}
The exact labels here are small and not important, but you  can see that the truth value means (right) are more similar to the MDN means (middle) than the simple RNN means (left). This confirms our suspicions in the earlier section that the MDN model, despite having more variance at each step, is usually regulated to be stable overall and correct major changes, preventing divergence, while the smooth RNN model may diverge. 

\subsection{Evaluation on OpenAI Baseline Learned Policies}
\begin{figure}[H]
    \includegraphics[width=\textwidth]{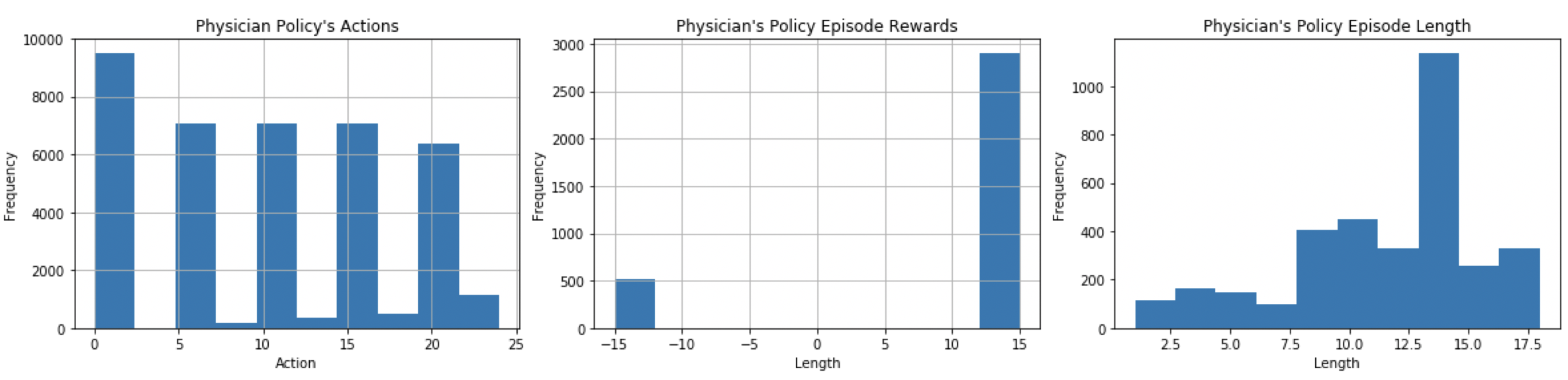}
    \caption{Physician's Policy}
    \label{doctor_policy}
\end{figure}

While simulated state  prediction results can be meaningful in helping us interpret different strategies, the end goal is to  learn a policy to improve patient outcomes, and thus we must evaluate our environment using a DQN agent algorithm. While there is no exact "label" or quantity to measure the clinical effectiveness of our learned policy (aside from clinical  validation), a qualitative comparison of the length, reward, and actions with those in the real dataset can give us a good assessment of how well our simulator models the treatment process. \ref{doctor_policy} shows the physician's policy distribution across actions, rewards and length, which we are comparing our policy to.

\begin{figure}[H]
\centering 
\includegraphics[width=1\textwidth]{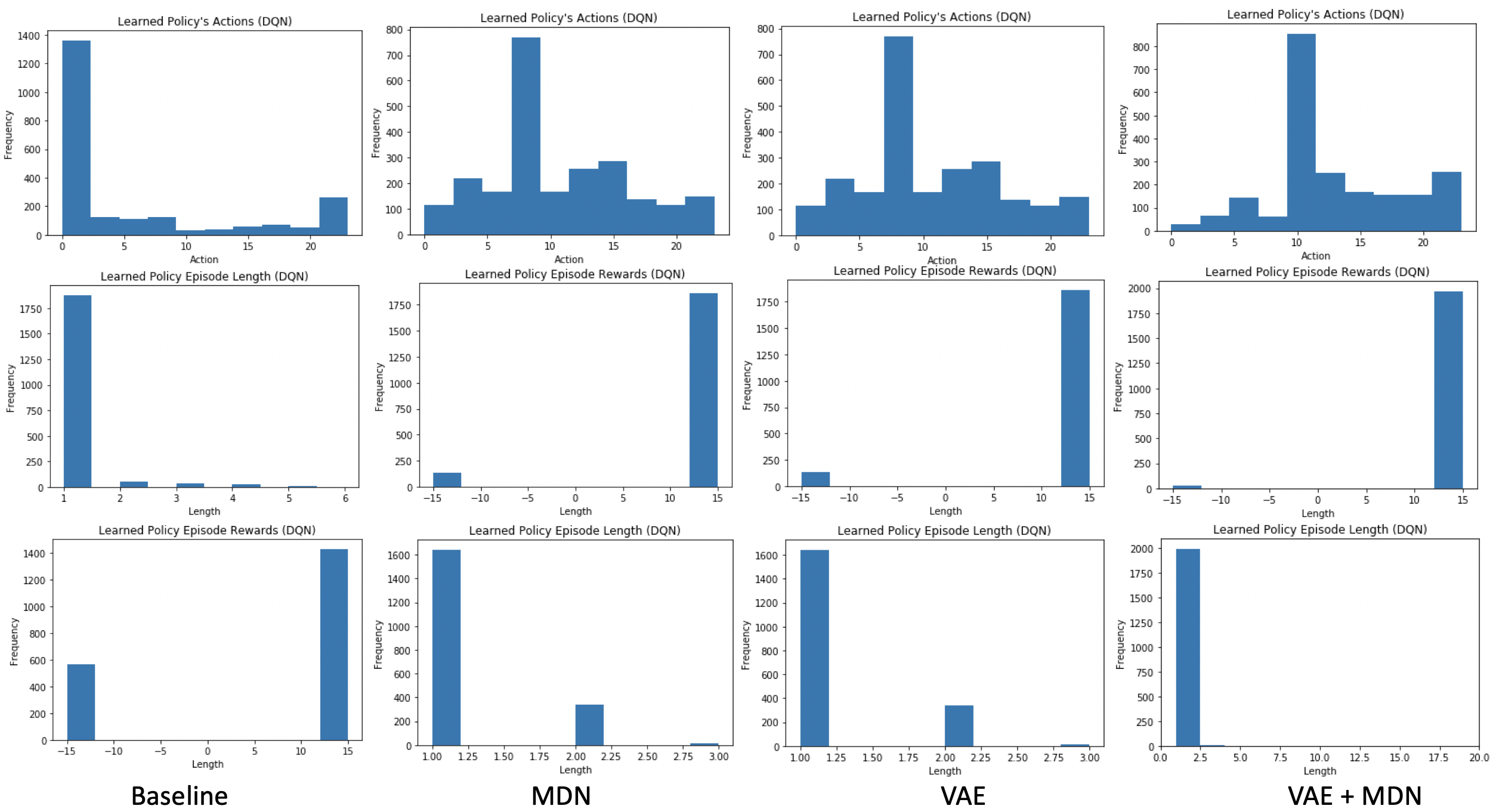}
\centering
\vspace*{-4mm}
\caption{Comparison of DQN learned policy for different networks using reward of 15.}
\label{15reward_plots}
\end{figure}

After replaying the physician's actions on our simulation environment, we compared the distribution of episode
length, rewards and actions between the real (shown above) and simulated worlds (our world-model). As mentioned earlier, we try using three different reward formulations. The results demonstrated in Figure \ref{15reward_plots}, which uses reward formulation (1) (all the reward at the end), seems to indicate an overly extreme simulation which causes the policy to cluster around one action  and a very short episode length. Rollout using the MDN-based simulator achieves a slightly more realistic state trajectory. The MDN model learns the distributions for each feature, giving us a more representative set of state features upon sampling. However, we suspect that the reason our learned policy is unrealistic compared to the physician’s policy is that the environment is overfitting to a small set of interventions and their positive outcomes in our dataset.  

We thus seek to address this issue by further enhancing the reward mechanism in the model to discourage extreme interventions. Using reward (2), as described earlier, we also get a model that prefers to take less extreme actions, which makes sense. However, the episode length still seems  rather unrealistic compared to that of the physicians. The  actions with  VAE+MDN are in fact even less  diverse than before, which  perhaps is to be expected given that the model wants to take as conservative actions as possible. In any case, this reward formulation also does not seem to give us the more diverse policy desired.

\begin{figure}[H]
\centering 
\includegraphics[width=1\textwidth]{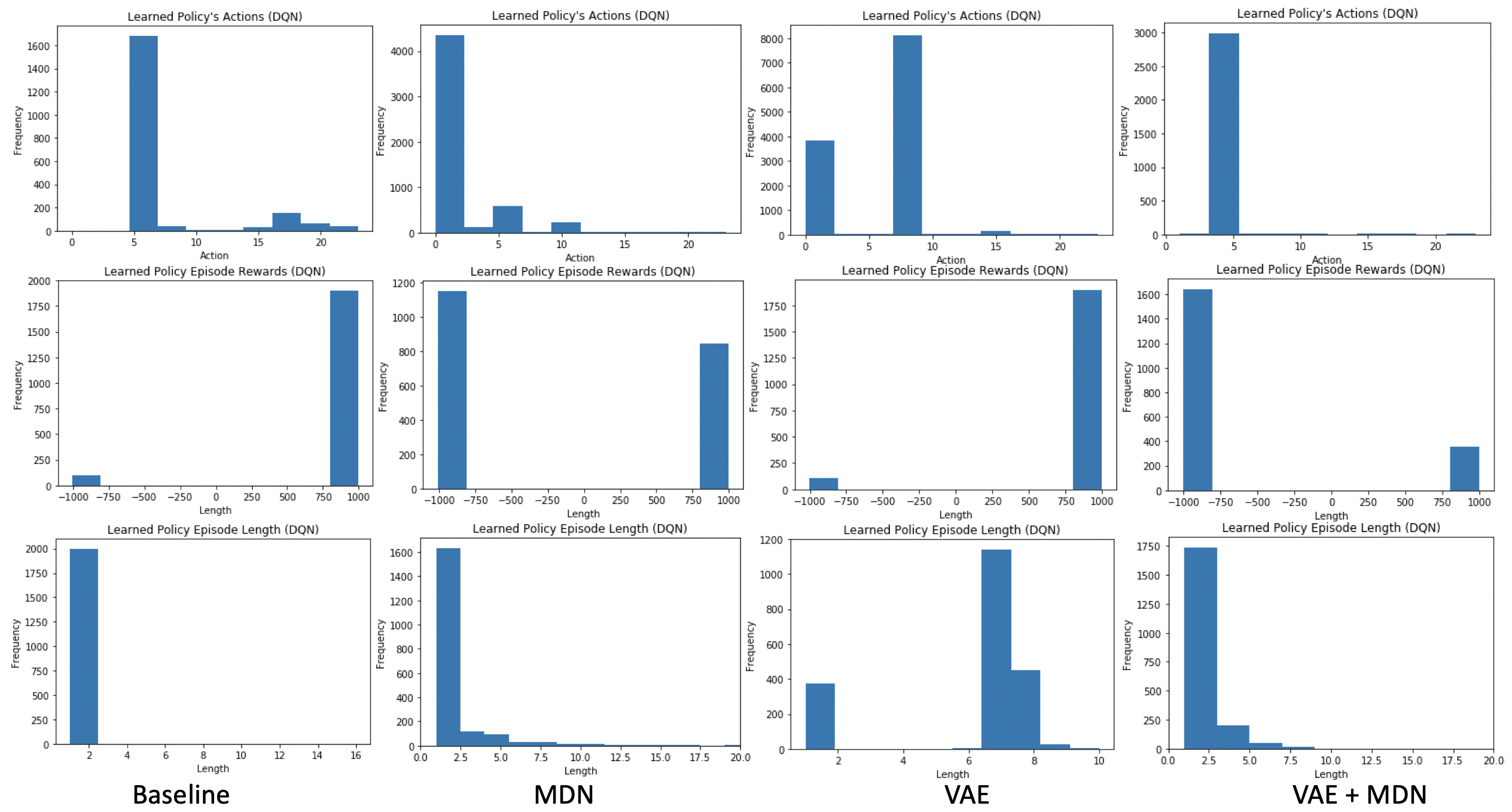}
\centering
\vspace*{-4mm}
\caption{Comparison of DQN learned policy for different networks using penalty of 1000.}
\label{1000penalty_plots}
\end{figure}

\begin{figure}[H]
\centering 
\includegraphics[width=0.7\textwidth]{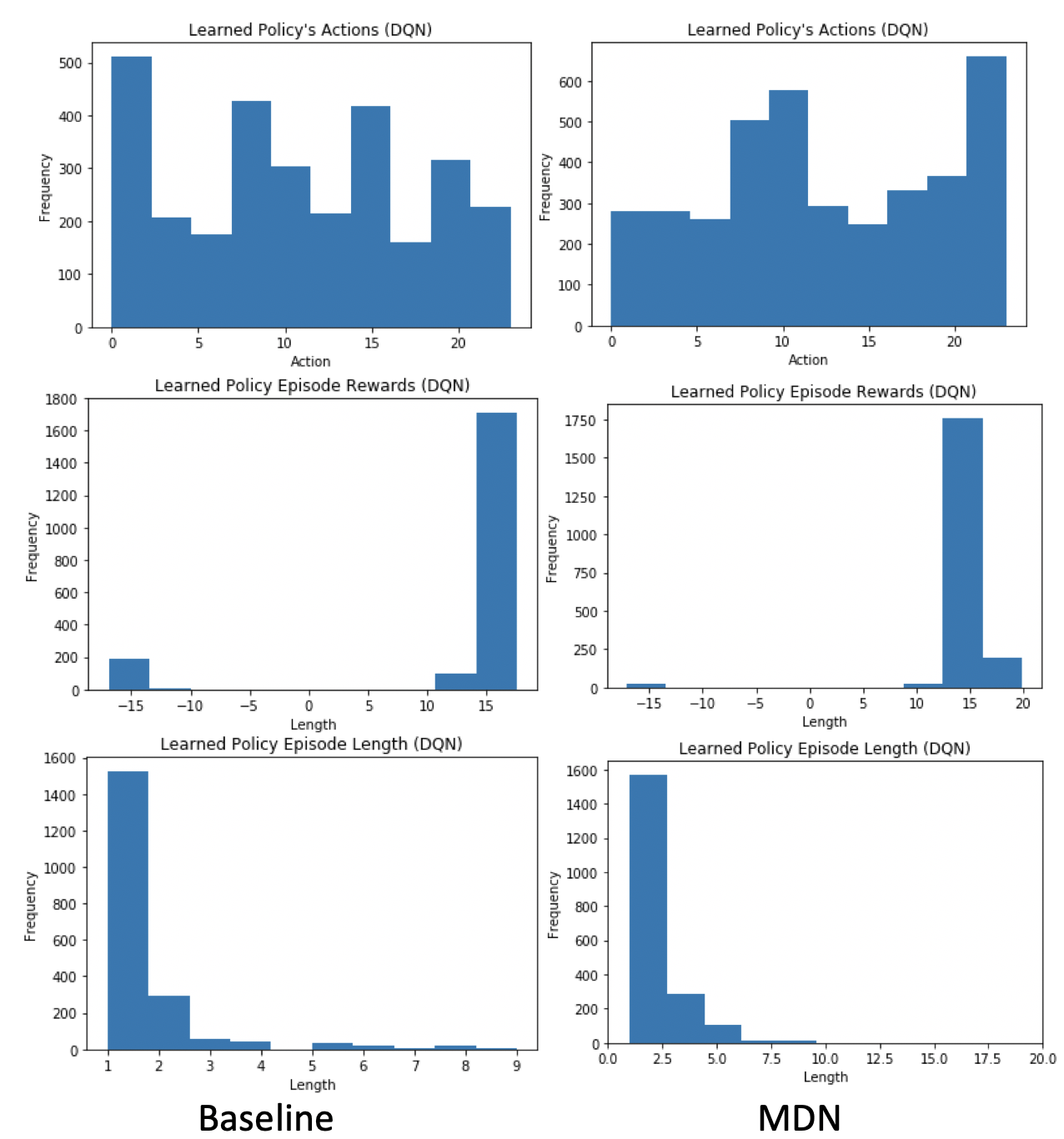}
\centering
\caption{Comparison of DQN learned policy for different networks using sofa reward of 15.}
\label{15sofa_plots}
\end{figure}

Finally, we use reward formulation (3), which uses changes in intermediate Lactate and SOFA levels to calculate rewards at each step. Here, as depicted in \ref{15sofa_plots}, we get a much more realistic distribution of actions. This could be due to that fact that the model has to choose an action  at each time step to optimize a specific value that will matter immediately, and therefore has an incentive to choose an optimal and specific action that works best for that state. In the other two reward formulations, the action needed was to optimize something at the end of the episode or was not time dependent, which caused the agent to predict the same action every time. 
However, we still have a very short episode length, which probably indicates that our termination model is overfit.

\vspace{-10 pt}
\section{Conclusion}
\subsection{Takeaways and Next Steps}
 Medical patients each have unique states and while it is an intractable problem to perfectly simulate the reactions of sepsis patients to medical interventions, we can try to improve our understanding of them through  more complex model systems. We have shown that two additional ways to learn uncertainty in our data, VAE and MDN, can better model the distribution of sepsis patient states than a simple RNN. We have also shown that we  can build a model on top of such simulator, and that various reward functions can be iterated  upon to model treatment trajectories. Future work is plenty in both of these angles, including optimizing the structure of the state/termination/outcome models, as  well as refinement of reward and uncertainty functions.
\subsection{Acknowledgement}
\vspace{-5 pt}
We are thankful for the mentorship of Peter Henderson (\textbf{phend@stanford.edu}), PhD Student at Stanford Computer Science Department, on this project.  We are also thankful for the feedback and support of our project mentor Benjamin Petit.

\vspace{-10 pt}
\section{Source Code}
The source code for this project can be access at \url{https://github.com/akiani/sepsisworldmodel221}. The packaged OpenAI Gym-based simulator can be accessed at \url{https://github.com/akiani/gym-sepsis}

\vspace{-10 pt}
\bibliographystyle{plainnat} 
{\footnotesize
\bibliography{bibfile.bib}}

\begin{thebibliography}{6}
\providecommand{\natexlab}[1]{#1}
\providecommand{\url}[1]{\texttt{#1}}
\expandafter\ifx\csname urlstyle\endcsname\relax
  \providecommand{\doi}[1]{doi: #1}\else
  \providecommand{\doi}{doi: \begingroup \urlstyle{rm}\Url}\fi

\bibitem[Ha and Schmidhuber(2018{\natexlab{a}})]{Ha2018}
David Ha and J{\"{u}}rgen Schmidhuber.
\newblock {World Models}.
\newblock 2018{\natexlab{a}}.
\newblock \doi{10.5281/zenodo.1207631}.
\newblock URL
  \url{http://arxiv.org/abs/1803.10122{\%}0Ahttp://dx.doi.org/10.5281/zenodo.1207631}.

\bibitem[Ha and Schmidhuber(2018{\natexlab{b}})]{world-model}
David Ha and J{\"{u}}rgen Schmidhuber.
\newblock World models.
\newblock \emph{CoRR}, abs/1803.10122, 2018{\natexlab{b}}.
\newblock URL \url{http://arxiv.org/abs/1803.10122}.

\bibitem[Johnson et~al.(2016)Johnson, Pollard, Shen, Lehman, Feng, Ghassemi,
  Moody, Szolovits, Celi, and Mark]{mimic-dataset}
A.~E. Johnson, T.~J. Pollard, L.~Shen, L.~W. Lehman, M.~Feng, M.~Ghassemi,
  B.~Moody, P.~Szolovits, L.~A. Celi, and R.~G. Mark.
\newblock {{M}{I}{M}{I}{C}-{I}{I}{I}, a freely accessible critical care
  database}.
\newblock \emph{Sci Data}, 3:\penalty0 160035, May 2016.

\bibitem[Liu et~al.(2018)Liu, Gottesman, Raghu, Komorowski, Faisal,
  Doshi{-}Velez, and Brunskill]{representation}
Yao Liu, Omer Gottesman, Aniruddh Raghu, Matthieu Komorowski, Aldo Faisal,
  Finale Doshi{-}Velez, and Emma Brunskill.
\newblock Representation balancing mdps for off-policy policy evaluation.
\newblock \emph{CoRR}, abs/1805.09044, 2018.
\newblock URL \url{http://arxiv.org/abs/1805.09044}.

\bibitem[Pinho and Costa(2018)]{Pinho}
Eduardo Pinho and Carlos Costa.
\newblock Unsupervised learning for concept detection in medical images: A
  comparative analysis.
\newblock \emph{Applied Sciences}, 8, 07 2018.
\newblock \doi{10.3390/app8081213}.

\bibitem[Raghu et~al.(2017)Raghu, Komorowski, Ahmed, Celi, Szolovits, and
  Ghassemi]{SepsisRL}
Aniruddh Raghu, Matthieu Komorowski, Imran Ahmed, Leo~A. Celi, Peter Szolovits,
  and Marzyeh Ghassemi.
\newblock Deep reinforcement learning for sepsis treatment.
\newblock \emph{CoRR}, abs/1711.09602, 2017.
\newblock URL \url{http://arxiv.org/abs/1711.09602}.

\end{thebibliography}

\vspace{-10 pt}
\section{Appendix}
\begin{figure}[H]
\centering 
\includegraphics[width=1\textwidth]{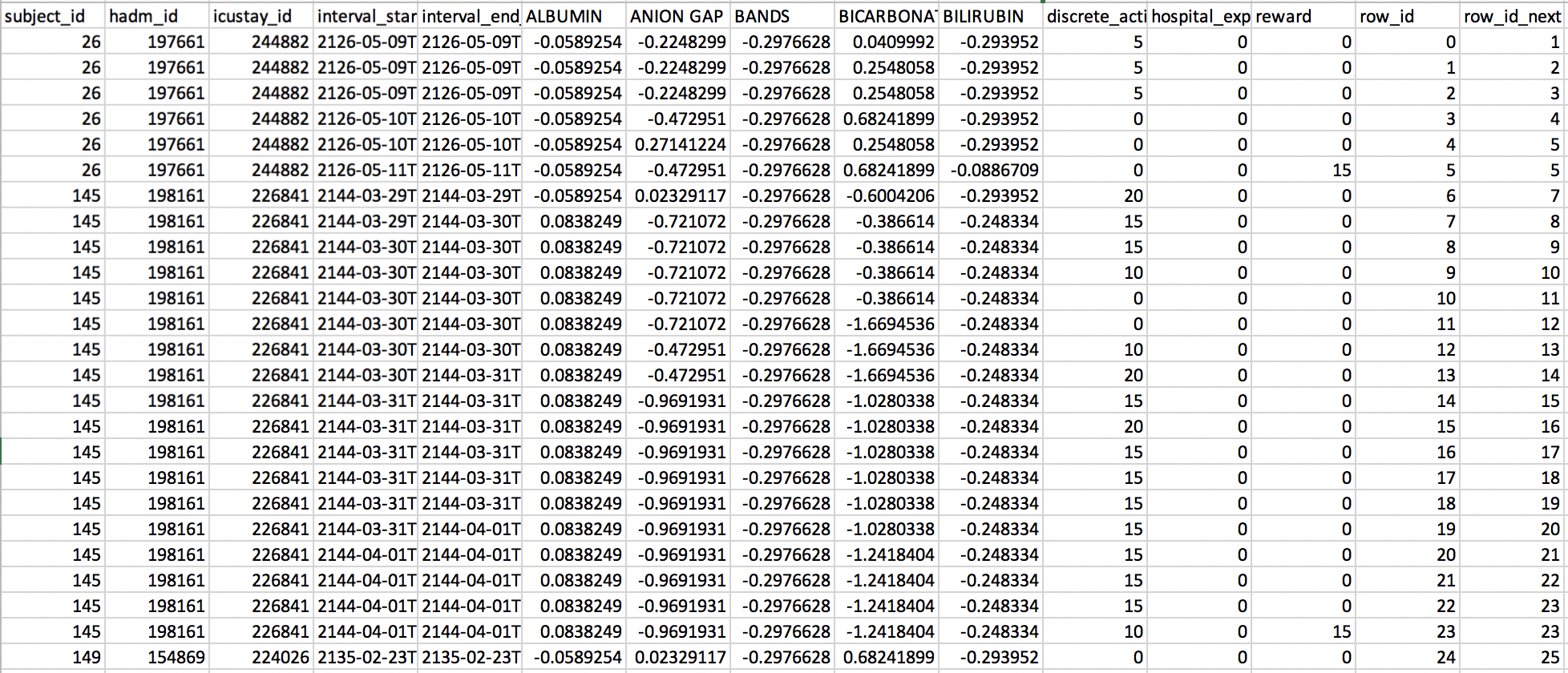}
	\caption{A sample of our preprocessed dataset. Notice how a single subject's datapoints over time are  associated, each with a set of features and an action taken to the next data point of that subject. The feature columns are condensed, there are dozens of more features.}
	\label{MDN}
\end{figure}

\end{document}